\documentclass[runningheads]{llncs}
\usepackage{graphicx}
\usepackage{algorithmic}
\usepackage{textcomp}
\usepackage{hyperref}
\usepackage{tikz}
\usepackage{xcolor}
\usepackage{tabularx}
\usepackage{multicol}
\usepackage{multirow}
\usepackage{booktabs} 
\usepackage{color}
\usepackage{hyperref}
\usepackage{listings}
\usepackage{color}
\usepackage{microtype}
\newcommand{\fakeparagraph}[1]{\smallskip\noindent\textbf{#1.}}
\newcommand*\circled[1]{\tikz[baseline=(char.base)]{
            \node[shape=circle,draw,inner sep=1pt] (char) {#1};}}

\begin{document}
\title{Towards Designing Computer Vision-based Explainable-AI Solution: A Use Case of Livestock Mart Industry\thanks{This is extended work of our demonstrations at ACM India Joint International Conference on Data Science \& Management of Data 8th ACM IKDD CODS and 26th COMAD, 2021. The authors thank Ciaran Feeney from MartEye, Prof. John Breslin from NUIG, Galway and Dr. Muhammad Intizar Ali from DCU, Ireland.}}
%
%
%
\author{Devam Dave\inst{1}, Het Naik\inst{1}, Smiti Singhal\inst{1}, Rudresh Dwivedi\inst{1} \and Pankesh Patel\inst{2}}
\authorrunning{D. Dave et al.}
%
\institute{School of Technology, Pandit Deendayal Energy University (PDEU), \\ Gandhinagar, India \\\email{\{devam.dce18,het.nce18,smiti.sce18,rudresh.dwivedi\}@sot.pdpu.ac.in} \and
AI Institute, University of South Carolina, Columbia, South Carolina, USA \\
\email{dr.pankesh.patel@gmail.com}}
\maketitle              
\begin{abstract}
The objective of an online Mart is to match buyers and sellers, to weigh animals and to oversee their sale. A reliable pricing method can be developed by ML models that can read through historical sales data. However, when AI models suggest or recommend a price, that in itself does not reveal too much (i.e., it acts like a black box) about the qualities and the abilities of an animal. An interested buyer would like to know more about the salient features of an animal before making the right choice based on his requirements. A model capable of explaining the different factors that impact the price point is essential for the needs of the market. It can also inspire confidence in buyers and sellers about the price point offered. To achieve these objectives, we have been working with the team at MartEye, a startup based in Portershed in Galway City, Ireland. Through this paper, we report our work-in-progress research towards building a smart video analytic platform, leveraging Explainable AI techniques.
\end{abstract}
\section{Introduction}
With the growth of AI and ML and deployment of advanced models in novel application domains, hitherto unasked questions and challenges come to the fore.  One such question relates to comparing the models to black boxes where it is not easy to understand their inner-workings, their algorithms and prediction reasoning seemingly stand unexplained.  This is also fuelled by the need for increasing efforts at removing bias where discriminatory features are found to be used by the model in the process of prediction. These two perceptions have led to increasing levels of distrust on ML/AI systems that are in use. Interpretability has also played an important role in applied ML over the years, but as black-box techniques such as DL take wings, it has assumed the form of concerns requiring urgent attention. As such, the field that attempts to fix these problems goes by the name of Explainable AI~(XAI)~\cite{doshivelez2017rigorous}.

An objective of this paper is to report our work-in-progress research and early results towards  building a smart video analytics platform, leveraging Explainable AI technology. We present a motivating scenario first and then report early results.

\section{Motivating Scenario}
The objective of an online Mart\footnote{\url{https://tinyurl.com/yyeytfc8}} is to match buyers and sellers, to weigh animals and to oversee their sale. Valuation of animals lack standards and transparency in the Livestock Marts Industry.  It is also a playground for agents who indulge in profiteering by selling or purchasing animals. The significant factors that should determine the price of the animals at the Mart are their lineage, physical attributes such as age, weight, and health. To ensure fair pricing of the animals, it is essential to understand data that determine their true value. Careful modelling and appropriate transparency consideration for valuation and pricing of animals should aid in stopping kickbacks, exorbitant commissions and price manipulations.

A reliable pricing method can be developed by ML models that can read through historical sales data. Moreover, it is possible for the visual recognition models of AI to recognize the features of animals and use that data for application of the pricing model. Collectively, it provides a valuable and unbiased tool for pricing of the animals at a Livestock Mart. \textit{However, when AI models suggest or recommend a price, that in itself does not reveal too much (i.e., it acts like a black box) about the qualities and the abilities of an animal. An interested buyer would like to know more about the salient features of an animal before making the right choice based on his requirements. } 

A model capable of explaining the different factors that impact the price point is essential for the needs of the market. It can also inspire confidence in buyers and sellers about the price point offered. To achieve these objectives, we have been working with the team at MartEye\footnote{\url{https://www.marteye.ie/}}, a startup based in Portershed in Galway City, Ireland. Through this paper, we report our work-in-progress research towards building a smart video analytic platform, leveraging Explainable AI techniques~\cite{doshivelez2017rigorous}.

\begin{figure*}[ht]
  \centering
  \includegraphics[width=\linewidth]{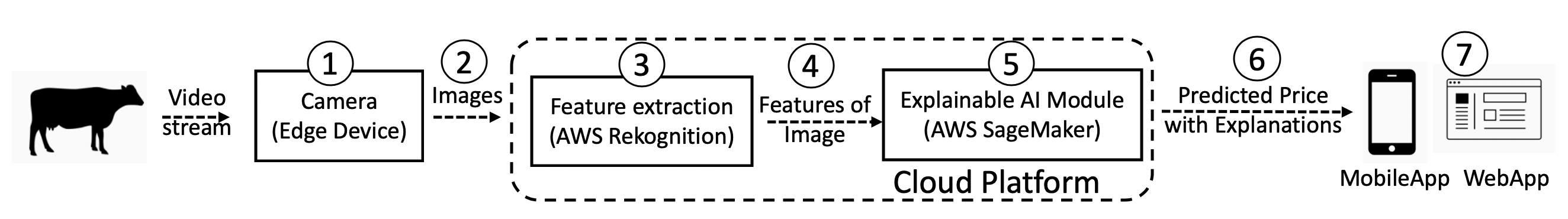}
  \caption{An Overview of Our Approach.}
  \label{fig:our-approach}
\end{figure*}

\section{Our approach and early results}

Figure~\ref{fig:our-approach} presents an overview of  our approach. 
In the following section, we present the software components and their implementation briefly:

\fakeparagraph{Edge Device} A real-time video stream is captured by a camera and pre-processed at the Edge device~(Circled~\circled{1} in Figure~\ref{fig:our-approach})~\cite{intizar:emse-01644333}~\cite{GYRARD2017305}~\cite{7460669}. The edge hosts frame sampling components that samples a frame off a live video stream from the attached camera and video pre-processing components to remove redundant and uninteresting parts of a video stream. Both the components are implemented using OpenCV library in Python.
The pre-processed video stream is packaged at the edge camera and transmitted~(Circled~\circled{2} in Figure~\ref{fig:our-approach}) to the AWS Cloud.

\fakeparagraph{Vision-based Feature Extraction}
A video stream is pre-processed at the Edge camera and transmitted to AWS.  AWS Kinesis ingests this video stream and it is analysed using AWS Rekognition, which lets developers build several computer vision capabilities on top of scalable and reliable Amazon infrastructure. We build feature extraction smart module using AWS Rekognition~(Circled~\circled{3} in Figure~\ref{fig:our-approach}).  It extracts several features from an animal image~(such as weight, age, height) by applying the classical and CNN-based vision techniques~\cite{SONG20184448}.


\fakeparagraph{ML-based Price Prediction} It takes features (extracted by the vision-based feature extraction module) and predicts the price of an animal.  We have trained a multivariate linear regression model using the historical dataset provided by the MartEye team. The team has collected large datasets from different marts across Ireland. The dataset contains 20+ features (e.g., \texttt{Price per kilo}~(PPK/Weight), \texttt{Weight of an animal}~(WT)). This model is trained by considering \texttt{the total price} of an animal as a target variable and the remaining features as independent variables. This model is implemented using AWS SageMaker, which provides the ability to build, train and deploy ML models on AWS.





\begin{figure}[ht]
  \centering
  \includegraphics[width=\linewidth]{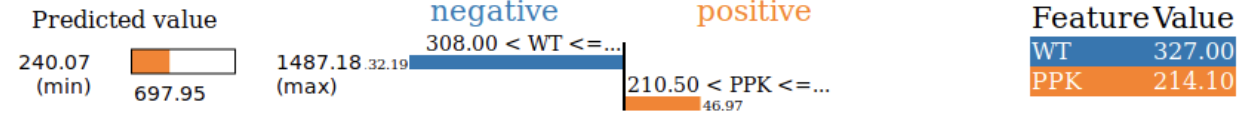}
  \caption{Model explanations generated by LIME.}
  \label{fig:lime}
\end{figure}

 The proposed approach generates price recommendations with explanations, describing how the recommended price is derived with explanations, instead of just predicting the price~(Circled~\circled{5} in Figure~\ref{fig:our-approach}). We employ several existing model interpretation techniques to generates insights from the multivariate linear regression model. Figure~\ref{fig:lime} illustrates the explanations, generated by LIME (Local Interpretable Model-agnostic Explanations) technique. The output of LIME is a list of explanations, considering the contribution of each feature  to the predicted value  of a dataset.

\fakeparagraph{Explainable AI Module} The proposed solution goes beyond prediction. It generates price recommendations with explanations, describing how the recommended price is derived at with explanations, instead of just predicting the price. We employ several existing model interpretation techniques to generate insights from the ML-based price prediction model.  Figure~\ref{fig:lime} illustrates the explanations generated by the LIME~(Local Interpretable Model-agnostic Explanations) technique, using the ML-based price prediction model. 

The output of Figure~\ref{fig:lime} is a list of explanations, considering the contribution of each feature to a predicted price. The left part  shows the range of a maximum~($1487.18$) and minimum~($240.07$) value, which is predicted by the ML-based price prediction module.  The middle part  shows the features~(i.e., $WT$ and $PPK$), which contribute the most in the predicted price of an animal. We can find that when the weight is in a range between $308.00<WT<=327.00$ it is contributing in a negative direction of the prediction.  We can also clearly see that when PPK is in a range $210.50<PPK<=214.10$ it is contributing to the positive side of the total price.  The right part shows the actual value of a particular feature (i.e., $Weight=327.00, PPK=214.10$). 







\section{Video Analytics for Weight estimation}

One of key objective of our approach is to estimate a weight value using video analytics. This section presents  our approach to estimate the weight.

\subsection{Weight estimation using Teachable Machine }
Teachable Machine is a web-based tool \cite{DBLP:journals/corr/abs-1905-07697} that is used for creating machine learning models in a simple way with no prerequisites required. Images, audios, and poses can be categorized using this tool, after which the model can be used in your own software. The main idea was to give a designated range of weights for the input image. We defined classes based on images for selected ranges and then gave an almost equal number of images for all the classes. The images used for testing included the cows with a familiar and unfamiliar background.
\begin{itemize}
    \item Phase 1: Single POV (fewer images with larger range)

In the first testing phase, we took 5 classes in the range of 200 KGs each, i.e 0-200, 201-400, etc. The images were taken from a single point of view, i.e side view. Approximately 30 images of each class were taken. The accuracy turned out to be approximately 92\%. In Figure~\ref{fig:gtm1}, the left screenshot below shows the image in which the model predicted the correct range of cow’s weight while the right image shows the one with the incorrect output predicted by the model.

\begin{figure}[!b]
  \centering
  \includegraphics[width=0.85\linewidth,height=9cm]{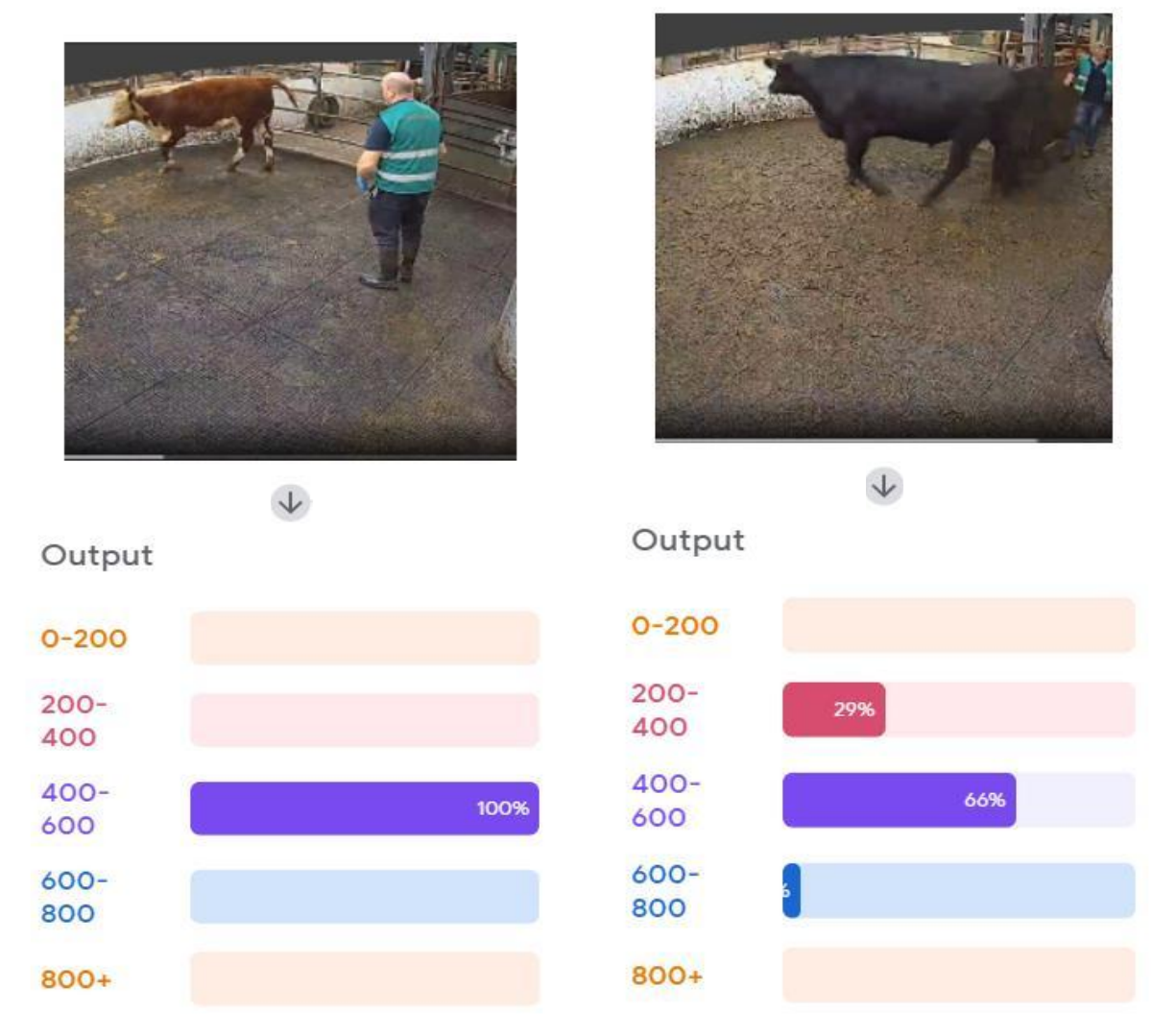}
  \caption{Prediction by teachable machine with 5 classes.}
  \label{fig:gtm1}
\end{figure}

\item Phase 2: Single POV (more images with smaller range)
 
In the second testing phase, we increased the number of classes to make the predicted value concise, keeping the view of the images the same. We took 8 classes in the range of 100 KGs each, i.e 0-100, 101-200, etc. The images were taken from a single point of view, i.e side view. Approximately 25 images of each class were taken. The accuracy turned out to be approximately 95\%. In Figure~\ref{fig:gtm2}, the left screenshot below shows the image in which the model predicted the correct range of cow’s weight i.e. 500-600, while the right one shows the one with the incorrect output predicted by the model.

\begin{figure}[t]
  \centering
  \includegraphics[width=0.85\linewidth,height=9cm]{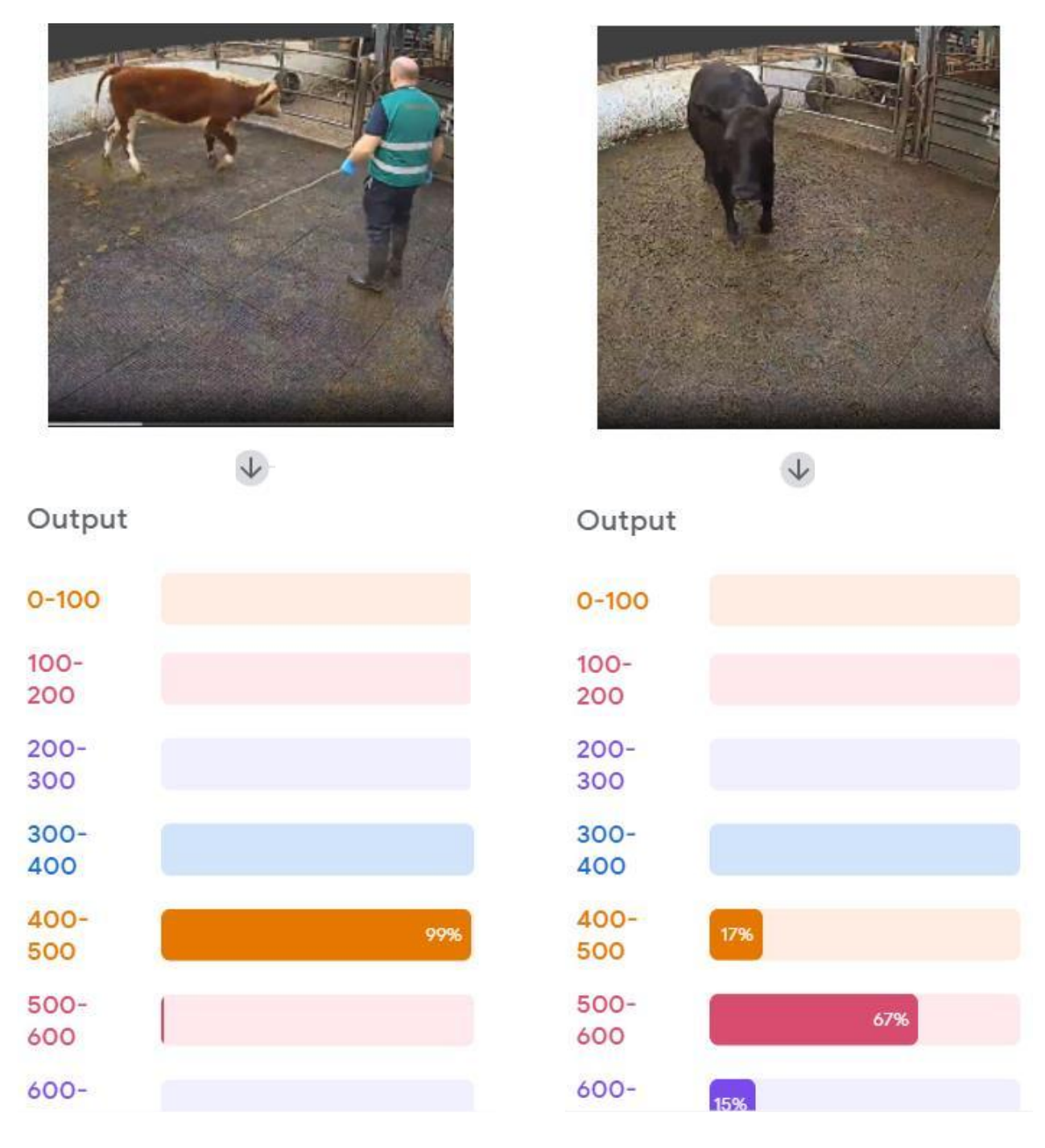}
  \caption{Prediction by teachable machine with 8 classes.}
  \label{fig:gtm2}
\end{figure}

\item Phase 3: Multiple POV (more images with smaller range)
	
In the third testing phase, we added different viewpoints of the cow ( i.e. front, back, and cross), keeping the number of classes and images the same as the previous phase, in order to make the predicted value concise. We took 8 classes in the range of 100 KGs each, i.e 0-100, 101-200, etc. Approximately 25 images of each class were taken. The accuracy turned out to be approximately 65\%.
Similar to the above 2 phases, Figure~\ref{fig:gtm3} shows the image in which the model predicted the correct range of cow’s weight i.e. 400-500, while the right one shows the one with the incorrect output predicted by the model.

\begin{figure}[ht]
  \centering
  \includegraphics[width=0.85\linewidth,height=9cm]{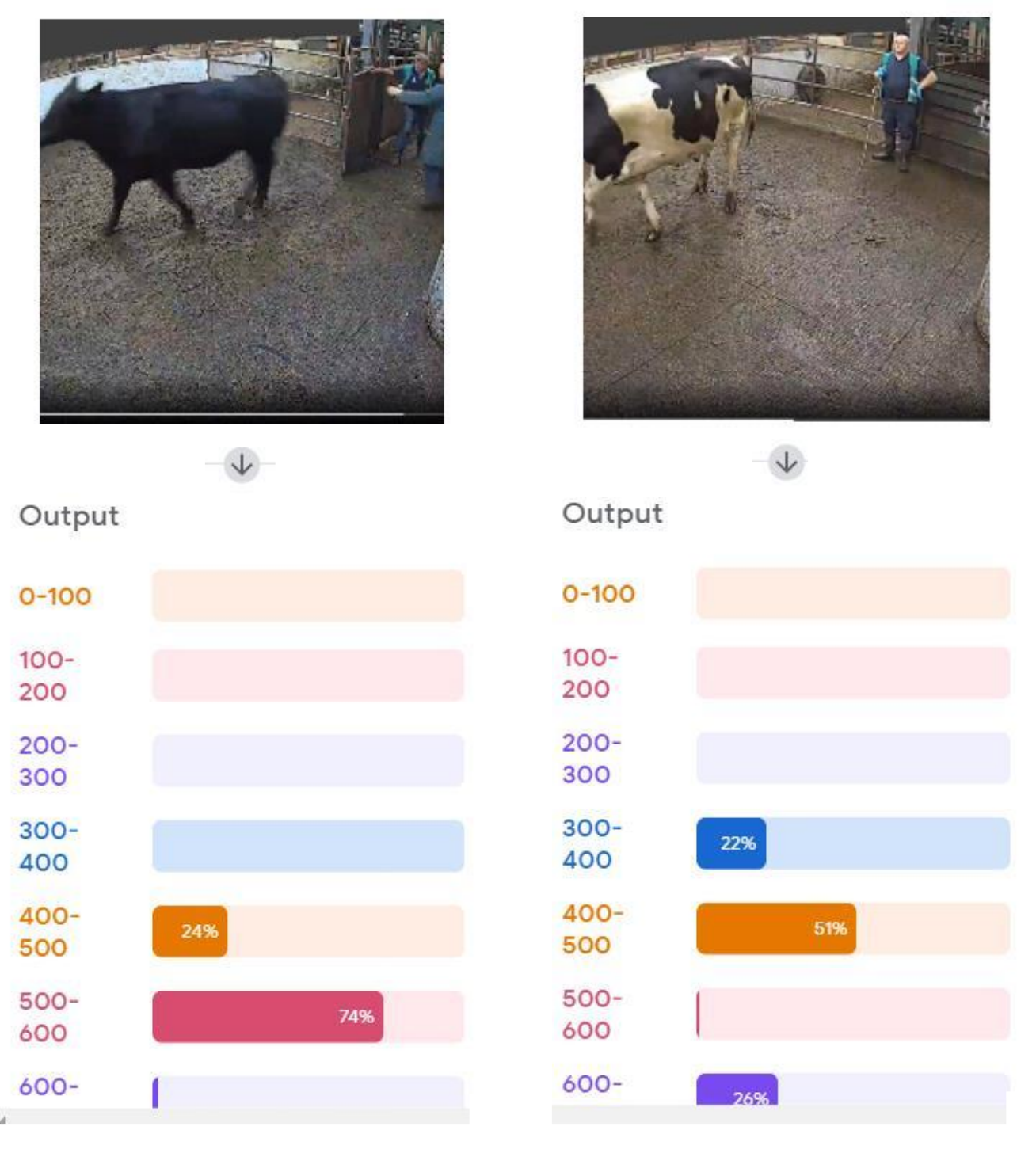}
  \caption{Prediction by teachable machine with different viewpoints.}
  \label{fig:gtm3}
\end{figure}
\end{itemize}

The Key observations from the method is that the accuracy varies based on training image's Point of View'. If the image of the cow is sideways, the model predicts almost the accurate weight, but when it is a front or back view, the model fails to predict the weight. Further, the color of the cow doesn't matter in this case, as the training data used contains a mix of images of cows of every breed (and color).

\subsection{Body Mass Index}

In this method, our main objective was to estimate the weight using a video frame that shows the front, side, and back view of the cows. The approach followed to find the weight was to capture an image of the face of the cow which was then given as input to the model to estimate the weight based on it. A ResNet based method \cite{kocabey2017face,royal-society,du2019techniques,angelov2020towards} was used for preparing the model and estimating the body weight from the input face, the reason being that ResNets not only have a significantly high accuracy of object classification, object detection and segmentation but also high training speed. Although the approach gives significantly good results for human beings, it fails to infer weight and deliver accurate results in our case.

The following limitations arise with this approach:
\begin{itemize}
    \item Unlike a human face that shows features such as age, gender, and a person’s skin being thick or thin depending on the facial fat, this is not the case with cows. All cows have similar facial features and they do not exhibit information that can be useful in weight estimation. Thus, the major issue is that the face won't be not enough for weight estimation.
    \item If we train by using the weight of each face of the cow, the model can get biased according to color, which will totally mislead the model and will give a set of vague predictions.
    \item The video quality provided is not good enough to capture the face in the video as zooming the video to capture the face does not give enough resolution to train the model on it.
\end{itemize}


\section{Acknowledgement}
This publication has emanated from research supported by grants from the European Union’s Horizon 2020 research and innovation programme under grant agreement number 847577 (SMART 4.0 Marie Sklodowska-Curie actions COFUND) and from SFI under grant numbers SFI/16/RC/3918 (Confirm) and SFI/16/RC/3835 (VistaMilk) cofunded by the European Regional Development Fund.

 \bibliographystyle{splncs04}
 \bibliography{mybibliography}

\end{document}